\documentclass[letterpaper,twocolumn,10pt]{article}
\usepackage{usenix2019_v3}

\usepackage{tikz}
\usepackage{amsmath}
\usepackage{algorithm}
\usepackage{algpseudocode}
\usepackage{multirow}
\usepackage{balance}
\usepackage{svg}
\usepackage{graphicx} 
\usepackage{algorithm}
\usepackage{algpseudocode}
\usepackage{booktabs}
\usepackage{cite}
\usepackage{amsmath,amssymb,amsfonts}
\usepackage{multirow}
\usepackage{hyperref}
\usepackage{balance}
[numbers]
\usepackage{stfloats}
\usepackage{xcolor}
\usepackage{svg}
\usepackage{verbatim}
\usepackage{array} 
\usepackage{caption} 
\usepackage{subcaption} 
\usepackage{subcaption}
\usepackage{amsmath}
\usepackage{pifont}
\usepackage{makecell}
\usepackage{balance}
\usepackage{authblk}

\usepackage{filecontents}

\begin{document}

\date{}

\title{\Large \bf FRAUD-RLA: A new reinforcement learning adversarial attack \\ against credit card fraud detection}

\author[1,2,3]{Daniele Lunghi}
\author[1]{Yannick Molinghen}
\author[2]{Alkis Simitsis}
\author[1,4,5]{Tom Lenaerts}
\author[1]{Gianluca Bontempi}

\affil[1]{Université Libre de Bruxelles}
\affil[2]{Athena Research Center}
\affil[3]{National and Kapodistrian University of Athens}
\affil[4]{Vrije Universiteit Brussel}
\affil[5]{Center for Human-Compatible AI, UC Berkeley}

\setcounter{Maxaffil}{0}
\renewcommand\Affilfont{\itshape\small}

\maketitle

\begin{abstract}
Adversarial attacks pose a significant threat to data-driven systems, and researchers have spent considerable resources studying them.
Despite its economic relevance, this trend largely overlooked the issue of credit card fraud detection. 
To address this gap, we propose a new threat model that demonstrates the limitations of existing attacks and highlights the necessity to investigate new approaches. We then design a new adversarial attack for credit card fraud detection, employing reinforcement learning to bypass classifiers. This attack, called FRAUD-RLA, is designed to maximize the attacker's reward by optimizing the exploration-exploitation tradeoff and working with significantly less required knowledge than competitors.
Our experiments, conducted on three different heterogeneous datasets and against two fraud detection systems, indicate that FRAUD-RLA is effective, even considering the severe limitations imposed by our threat model. 

\end{abstract}

\section{Introduction}


In the first half of $2023$, credit card payments in the euro area accounted for over $50$ trillion euros~\cite{ECB_Report}. 
Given the size of the domain and its economic relevance,  ensuring the robustness of payment systems is paramount, and vast resources are continuously being spent on improving the quality of credit card fraud detection systems. In particular, the role played by machine learning can hardly be overestimated~\cite{abdallah2016fraud}. 
Research on fraud detection, however, has focused on the statistical properties of frauds~\cite{dal2017credit, van2015apate}, and proper assessments of the risk posed by adaptive fraudsters' strategies are lacking in the literature. 
In most domains, robustness is assessed through \textit{adversarial attacks}~\cite{song2020mab, pintor2023imagenet}, i.e., attacks designed against machine learning models. 
Although many such attacks have been created over the past decade (see Section \ref{sec:SOTA}), the majority of them focus on image recognition~\cite{song2020mab, carminati2020evasion}, posing a significant challenge in terms of their generalization to credit card fraud detection. 

Notably, only a handful of works have studied adversarial machine learning in the context of credit card fraud detection.
The main works~\cite{carminati2018security, carminati2020evasion} attack the same realistic fraud detection engine called BankSealer~\cite{carminati2015banksealer}. In both works, the authors rightfully consider domain-specific challenges generally absent in other adversarial works, such as the intricate feature engineering process performed in fraud detection.
However, they operate under the assumption that fraudsters can access the customers' transaction history. 
As the authors point out, this may be achieved through the introduction of malware into the victim's devices. However, this considerably increases the difficulty of performing any attack, as fraudsters must first compromise the customer's device and observe past transaction history, which constitutes a significantly more complex undertaking than stealing or cloning a card. 
This may limit the scalability of such attacks and, therefore, their capacity to compromise the overall system security.


Our work aims to fill the gap in the literature between the fields of credit card fraud detection and adversarial machine learning. To achieve this, our work makes two contributions. 
First, it provides a novel formulation for adversarial attacks against fraud detection systems. 
\textcolor{black}{
Past works have discussed the lack of human supervision~\cite{Lunghi2024Assessing} and the significance of aggregated features~\cite{lunghi2023adversarial}. These works focused on how constraints impacted existing adversarial attacks against fraud detection but did not lead to the development of a systemic analysis of the threat posed.
In this work, we overcome such limitations by proposing a new threat model (Section~\ref{sec:Threat}) specifically designed to tailor the main characteristics of credit card fraud detection. The definition of the primary challenges and advantages that fraudsters encounter compared to our domains results in the design of \autoref{tab:AttacksUsability}, which demonstrates how existing approaches fail to cope with unknown features, restricted access to the fraud detection engine, and the inability to exploit the absence of human investigation.
}

Our main contribution is a novel adversarial attack against credit card fraud detection systems based on Reinforcement Learning~\cite[RL]{sutton_reinforcement_2018}. 
We model the problem of generating fraudulent transactions not detectable by the classifier as an RL problem where the agent is assigned the task of crafting the fraudulent transactions. In that context, the agent is rewarded when it successfully fools the classifier.
We identify Proximal Policy Optimization~\cite[PPO]{SchulmanWDRK17-ppo} as a suitable RL algorithm, and we use it to create a new attack named FRAUD-RLA (Reinforcement Learning Attack).  
Our analysis and experiments show that FRAUD-RLA is the most effective attack against credit card fraud detection systems. Unlike previous works, FRAUD-RLA does not require the same degree of access to the model or the users' transaction history. Moreover, using RL allows us to explicitly model the tradeoff between the quality of the attack pattern found and the time required to find it. As we discuss in Section \ref{sec:Threat}, this tradeoff, also known in RL literature as the exploration-exploitation tradeoff~\cite{macready1998bandit}, is an essential feature of attacking fraud detection systems.
To assess the effectiveness of  FRAUD-RLA, we test our method against fraud detection engines on real and synthetic datasets. Our experiments, reported in Section \ref{sec:results}, show the effectiveness of the reinforcement learning approach.

\textbf{Outline}.
\textcolor{black}{
The rest of the paper is structured as follows.  Section~\ref{sec:background} provides the background for this work, explaining the threat model we use and introducing the main attack approaches in the literature.
Section~\ref{sec:FrauRLA} introduces FRAUD-RLA and explains its functioning.
Section~\ref{sec:Experiments Design} discusses the experimental assessment of our method, and Section \ref{sec:Conclusions} concludes the paper.
}

\section{Background} \label{sec:background}

This Section is composed of two main parts. First, we define the threat model describing the target, i.e., the fraud detection system and the attackers. Then, we present existing attack approaches and compare them against this model.

\subsection{Threat model} \label{sec:Threat}

\subsubsection{Credit Card Fraud Detection}
In online payment systems, cardholders use their card to access a terminal and perform transactions. Essentially, a transaction consists of an amount paid to a merchant by a cardholder at a given time~\cite{leborgne2022fraud}. Specific features, such as the card brand and the merchant country, characterize cards and terminals.
We call  \textit{raw features}  these features and the ones directly chosen by the user, such as the transaction amount.
Furthermore, cards and terminals have an identifier, allowing the fraud detection system to aggregate transactions based on the card or terminal used in the process. These aggregations are then used to generate a set of features called \textit{aggregated features}. Such transactions can be built around the customer~\cite{carminati2015banksealer}  or terminal~\cite{lunghi2023adversary} history.
Raw and aggregated features are then used to train a fraud detection system. This system is built on a composition of human-made rules and machine learning~\cite{dal2017credit, carcillo2018scarff}, which combine to signal the most likely frauds to the investigators.

\subsubsection{Attackers}

Discussing attackers' goals, knowledge, and capabilities is crucial to designing realistic attacks, which can be used for robustness assessments~\cite{joseph2019adversarial}.

In credit card fraud detection, the attackers' goal is straightforward: deceive the fraud detection system into misclassifying their frauds as genuine transactions.
However, two main details differentiate this field from other domains.
First, fraudsters can access a limited set of cards, which may be blocked over time. Hence, they need to maximize the number of successful frauds from the beginning through an exploration-exploitation tradeoff~\cite{macready1998bandit}.
This is particularly true as fraud detection data are subject to concept drift~\cite{dal2017credit}, forcing the classifiers to adapt to be up-to-date with reality.
While these transformations may be too slow to effectively block attacks, given the speed at which they could be performed, fraudsters may discover that the attack patterns they found do not work once the model has been updated. 
Furthermore, adversarial attacks generally work under the implicit assumption that humans may observe the attacks and detect anomalies, forcing the attackers' perturbation to be imperceptible~\cite{luo2018towards,khamaiseh2022adversarial} to bypass such a human check.
In fraud detection, imperceptibility should not be part of the attackers' goal. Since humans observe only a few suspicious transactions, bypassing the automatic checks is enough to perform a successful attack. 

Defining attackers' knowledge is more complex. 
A common principle in computer security literature is "no security through obscurity"~\cite{scarfone2008guide}, meaning that it is better to assume secret information could always fall into the wrong hands. Therefore, considering the attackers' perfect system knowledge is generally better. Credit card fraud detection, however, presents a peculiar situation. While it is true that a skillful attacker may, in principle, retrieve any information (for instance, through social engineering~\cite{salahdine2019social}), such attacks are unlikely to be scalable enough to pose a significant threat to fraud detection systems. 
For this reason, we assume that attackers may know the principles around which the system is built, like its features engineering process, but not the weights of the trained classifier, which change over time.

\begin{figure*}
    \centering
    \includegraphics[scale=0.4]{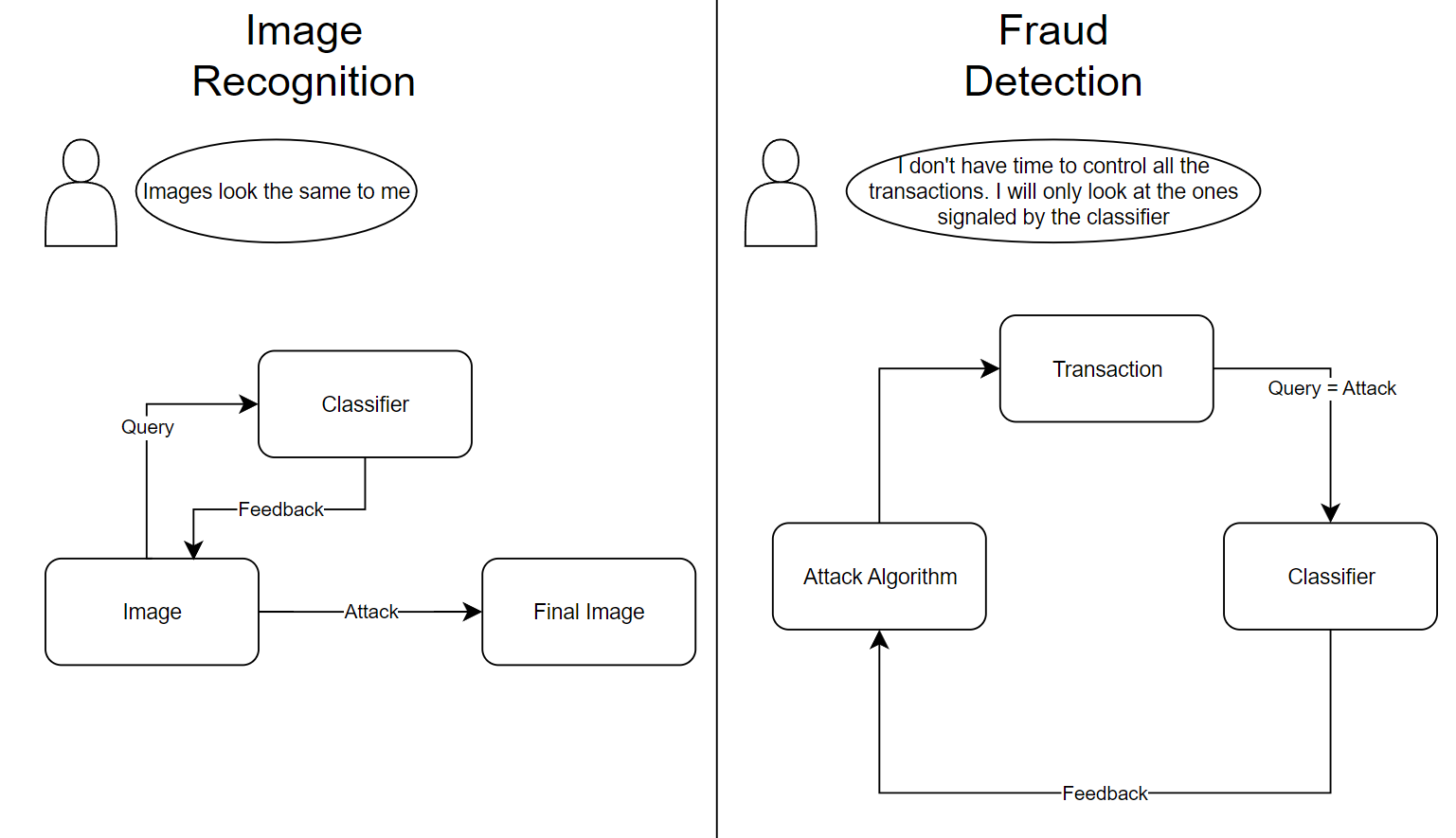}
    \caption{Visual representation of attacks in image recognition (left) and fraud detection (right)}    \label{fig:Visual_Fraud_rest}
\end{figure*}

A second form of knowledge, specific to this application, is \textit{data knowledge}, not to be confused with the \textit{features knowledge} described in~\cite{suciu2018does}.
While \textit{features knowledge} focuses on knowing the features engineering technique applied by the classifier, it still assumes that the attacker has complete knowledge of the observation location in the original data space.
In credit card fraud detection, however, each transaction is also evaluated based on aggregated features. To know these, fraudsters must reconstruct the card's and terminal's transaction history, which is challenging. Skillful attackers may achieve this through the use of malware previously injected into the devices~\cite{carminati2020evasion}, but this poses a strong requirement on the attackers' side and may limit the number of attacks they can perform.
For this reason, we work under the more general assumption that attackers know the feature transformation and the general model characteristics and may know some fixed card features, but do not know the aggregate features.
Previous literature corroborates this hypothesis, as most fraudsters seem not to be influenced by the cardholder's behavior before their first fraud with a new stolen card\cite{lunghi2023adversary}.

Let us now define attackers' capabilities. First, they are able to decide the value of transaction features, such as the amount. 
Instead, they cannot modify fixed card features. They may, in principle, find values that allow them to adjust as they want the aggregated features~\cite{carminati2020evasion}. However, this would require a complete knowledge of the feature generation process and the previous transactions performed with the used card and terminal, which they do not generally have.

A graphical representation of the differences between fraud detection and other adversarial attacks in terms of indistinguishability and the query loop is illustrated in \autoref{fig:Visual_Fraud_rest}, highlighting differences between the two settings. First, image recognition attacks are designed to be indistinguishable by humans, but investigators can only observe the most likely frauds in fraud detection. The attacks loop is also different: while standard adversarial attacks perform multiple queries before submitting each attack, fraudsters are evaluated over each transaction, with no difference between query and attack time.

\subsection{Adversarial Attacks} \label{sec:SOTA}

Let us use the defined threat model to design a taxonomy of existing adversarial attacks based on their applicability against credit card fraud detection. Coherently with our threat model, we focus here on attacks that do not require knowing the target classifier, also called \textit{Black Box attacks}~\cite{joseph2019adversarial}.
First, based on their approach, we group these attacks into four main groups. Then, how some of these attacks have been used in the context of credit card fraud detection.



\subsubsection{Taxonomy}

\paragraph{Image recognition / Query-based} Query-based attacks acquire information through queries, i.e., attempted attacks, and use feedback provided by the target classifier to improve their attack quality. Most attacks designed against image recognition systems fall into this category~\cite{chen2017zoo, brendel2017decision,chen2020hopskipjumpattack}.   
For instance, Boundary Attack~\cite{brendel2017decision} first samples the data space to find a valid attack and then gets closer to the original observation by moving along the decision boundary of the model. Notably, this requires continuously estimating the decision boundary locally, which can only be obtained by sending observations to the target model and observing the associated decisions.
Applying these attacks to credit card fraud detection is exceptionally challenging. First, they have no mechanism to optimize the exploration-exploitation tradeoff, as the locality of their attacks means a constant number of queries for each attack. Similarly, image recognition attacks are designed to be imperceptible to human eyes, making them incredibly inefficient when such constraint is irrelevant~\cite{Lunghi2024Assessing}. Finally, they assume full data knowledge to estimate the local boundaries.

\paragraph{Surrogate Model} The idea behind surrogate models~\cite{demontis2019adversarial, debicha2023adv} is relatively straightforward. Since querying the target classifier may not always be feasible or efficient, if the attacker can access a labeled training set, they can train a surrogate model whose properties should mimic the ones of the target. Attackers then test their attack on the resulting classifier, also called surrogate, and use attacks transferability~\cite{demontis2019adversarial} to break the target classifier.
Fraud detection datasets, however, are realistically challenging for an attacker to obtain, especially considering these attacks generally require a labeled dataset. Their reduced availability makes model surrogate attacks extremely hard to apply.

\paragraph{Mimicry}
The idea behind Mimicry is that if an attacker has access to some data from a class, they might craft attacks that mimic those data so that the classifier recognizes them as belonging to that class~\cite{wagner2000intrusion}. Attackers mimic the behavior of a particular user while satisfying a set of domain constraints~\cite{carminati2018security}.
The key advantage compared to Surrogate Model attacks is that Mimicry attacks only need genuine data. 
Credit card fraud detection data are highly unbalanced, meaning that any random sample of data is likely to be composed mainly of genuine data. As such, an unsupervised dataset is likely composed primarily of genuine transactions~\cite{dal2014learned} and would be a good choice to train Mimicry.

\paragraph{Reinforcement Learning}
Reinforcement learning has been used as an adversarial attack tool in cross-site scripting~\cite{fang2019rlxss} and malware detection~\cite{anderson2018learning, tsingenopoulos2022adaptive}.
These works, however, are domain-specific, and their application to credit card fraud detection leaves many unanswered questions.
First, these attacks are designed to modify malicious code while maintaining its functionality. Therefore, they require a starting point (for instance, the original malware) and a set of possible transformations to perform on it. 
In credit card fraud detection, however, attackers can choose the values of their features freely, without a starting point to modify. Moreover, these attacks are designed with the assumption that once an attack is found, the attacker can reproduce it to infect any number of devices employing the same detection system. This is not true in fraud detection, where attackers have a limited number of cards and have a strong incentive to find working attacking patterns in the shortest possible time.

\subsubsection{Attacks on Credit Card Fraud Detection}

Previous works describing attacks against card fraud detection (\textit{Fraud attacks} in \autoref{tab:AttacksUsability}) assume that fraudsters can inject the victim's device with a banking Trojan, which allows them to perform frauds and, crucially, observe the transactions previously performed using the card~\cite{carminati2018security,carminati2020evasion}. 
These works use different approaches to solve the problem.
First, fraudsters can use a previously acquired dataset of genuine transactions to build a model for genuine users' behavior and then generate sequences of frauds that avoid detection by blending into genuine transactions' patterns~\cite{carminati2018security}.
Alternatively, if the attacker has access to a set of transactions resembling those used in the training set of the algorithm, they can train a substitute model, called Oracle, and use it to craft their frauds~\cite{carminati2018security}. Interestingly, such a training set can be old or belong to a different institution if the patterns behind the data are close enough to allow for attack transferability. The attacker then generates raw transactions; the fraud detection engine aggregates and evaluates them.

Both attacks can be performed in a Black-Box setting, and using a training set to model the target classifier before starting the attacks implicitly solves the exploration-exploitation tradeoff. The flip side is that they require data to build a surrogate training set, which may be hard to acquire for most attackers and can be a costly operation.
More importantly, attacking a customer requires injecting their device with a banking Trojan, which is generally a much stronger assumption than simply cloning or stealing the card. This may limit the number of attackers capable of implementing such a strategy, reducing the threat such attacks pose to fraud detection engines, as they cannot be performed with the same frequency of attacks with fewer requirements. Finally, these attacks assume that classifiers use only raw and customer-based features, as is the case for \cite{carminati2015banksealer}. However, various works in the literature also use terminal-based aggregations~\cite{jha2012employing, carcillo2018scarff}. To attack them using these methods, fraudsters would also need to control the terminal, further increasing the difficulty of performing these attacks.


\begin{table*}[]
\resizebox{\textwidth}{!}{%
\begin{tabular}{|l|l|l|l|l|l|l|}
\hline
\textbf{Attacks} &
  \textbf{Black Box} &
  \textbf{Imperceptibility} &
  \textbf{Starting point} &
  \textbf{\begin{tabular}[c]{@{}l@{}}Data\\  Knowledge\end{tabular}} &
  \textbf{\begin{tabular}[c]{@{}l@{}}Exploration-Exploitation\\ Tradeoff\end{tabular}} &
  \textbf{\begin{tabular}[c]{@{}l@{}}Training Set\\ Access\end{tabular}} \\ \hline
Query-Based     & YES & NO        & NO        & NO        & NO        & YES       \\ \hline
Surrogate Model & YES & Partially & Partially & Partially & YES & NO        \\ \hline
Mimicry         & YES & YES & Partially & Partially & YES & Partially \\ \hline
Fraud Attacks   & YES & YES       & YES       & NO        & YES       & NO/Partially       \\ \hline
FRAUD-RLA       & YES & YES       & YES       & YES       & YES       & YES       \\ \hline
\end{tabular}%
}

\caption{Compliance of prominent adversarial attack families to fraud detection constraints. Attacks can completely fulfill the requirement (YES), be designed in a way that makes it exceptionally challenging to tackle (NO), or partially fulfill the requirement because their design does not explicitly make it difficult to do so. With FRAUD-RLA we refer to the attack proposed in this paper. }
\label{tab:AttacksUsability}
\end{table*}

\section{FRAUD-RLA: Reinforcement Learning Against Credit Card Fraud Detection} \label{sec:FrauRLA}

In this Section we present our main contribution, a new attack against credit card fraud detection called FRAUD-RLA. The design of FRAUD-RLA involves three core elements: 
\begin{itemize}
    \item \textbf{Problem Formulation}, where we formalize the concepts of transaction and fraud detection engine and the attacker's task as discussed in Section \ref{sec:Threat}.
    \item  \textbf{Task definition}, where we reformulate the problem as a single-step Partially Observable Markov Decision Process (POMDP)~\cite{sutton_reinforcement_2018, oliehoek_concise_2016}, allowing us to employ reinforcement learning algorithms.
    \item  \textbf{Solution design}, where we describe how we find the solution through Proximal Policy Optimization~\cite[PPO]{SchulmanWDRK17-ppo}, a gradient-based technique capable of online learning the best attack policy.
\end{itemize}

\begin{algorithm}
\caption{FRAUD-RLA to maximize the sum of successful frauds (total reward) over training time}\label{alg:fraud_detection}
\begin{algorithmic}[1]
\Require Time budget $t_{\max}$, Fraud Detection Engine $f$, Set of fraudulent transactions $X$, environment $M$

\State Initialize $PPO$ parameters $\theta$
\State $t \gets 0$
\State $total\_reward \gets 0$

\While{$t < t_{\max}$}
    \State Receive $x_k, x_u$ from $M$ \label{line:getfeatures}
    \State $\boldsymbol{\mu, \Sigma}  \gets PPO.compute\_means\_and\_covariance(x_k)$ \label{line:compute_meansCov}
    \State $x_c \sim \mathcal{N}(\boldsymbol{\mu, \Sigma})$ \label{line:bestControl}
    \State $reward \gets  1 - f(x_c, x_k, x_u)$ \label{line:collect-rewardPPO}
    \State $PPO.store(x_k, x_c, reward)$ \label{line:update1}
    \State Update $\theta$ according to PPO loss \label{line:update2}
    \State $t \gets t + 1$
    \State $total\_reward \gets total\_reward + reward$
\EndWhile
\end{algorithmic}
\label{alg:FRAUD-RLA}
\end{algorithm}

\subsection{Problem Formulation}

To formulate the problem definition of the attack, we define the attack's objects (the transactions), the target (the classifier), and the attacker.
First, we define a transaction as a triplet: 
\begin{equation}
   x = \left< x_c, x_k, x_u \right>
\end{equation}
where $x_c \in \mathbb{R}^C$ are the features \emph{controllable} by the attacker, such as the amount of the transaction, $x_k \in \mathbb{R}^K$ are the features \emph{known} to the attacker, such as the card number, and $x_u\in \mathbb{R}^U$ are the features \emph{unknown} to the attacker, such as the history of previous transactions. We note $C$ the number 
 of controllable features, $U$ the number of unknown features, and $K$ the number of known features. 

Let us now consider a fraud detection engine called $f$, which takes a transaction $x$ as input and returns a decision $f(x) = \{0, 1\}$ on whether to block the transaction (and the associated card). This fraud detection engine is internally composed of a mix of rule-based and data-driven classification algorithms called $f_r$ and $f_d$, respectively. A transaction is considered genuine if accepted by both classifiers, i.e., $f(x) = 0$ if $f_r(x) = 0$ and $f_d(x) = 0$. \textcolor{black}{Such a classifier is inspired by previous works in the literature (e.g.,~\cite{dal2017credit, carcillo2018scarff})}. 
In the decision-making process, the attacker observes $x_k$ and then determines the values of $x_c$. Then, they submit $x_c$ to the fraud detection engine, which evaluates the composed transaction $x = \left< x_c, x_k, x_u \right>$, and returns the associated label $f(x)$.
The fraudster can then perform another transaction, which will, in principle, have different values of  $x_k$ and $x_u$. To model the limited number of frauds an attacker can perform, this operation is repeated for $t_{max}$ rounds.

\subsection{RL environment}
\label{sec:environment}

We model the environment as a single-step Partially Observable Markov Decision Process (POMDP)~\cite{sutton_reinforcement_2018, oliehoek_concise_2016}, denoted as $M$. Formally, this can be written as: $M = \left<S, O, A, T, R, \Omega\right>$, where $S$ is state space, $O$ is the observation space, $A$ is action space, $T$ is the transition function, $R$ is the reward function, and $\Omega$ is the observation function. Each of these components is defined as follows.

\begin{itemize}
    \item $S$ is a continuous space of size $\mathbb{R}^{U + K + C}$, i.e. every possible transaction.
    \item $O$ is the continuous space set of possible observations of size $\mathbb{R}^{K}$, i.e. every possible known features.
    \item $A$ is a continuous action space of size $\mathbb{R}^C$.
    \item $T: S \times A \rightarrow S$ is a deterministic transition function that maps each state-action $(s, a)$ pair to a next state $s'$. Note that $s$ is always an initial state and that $s'$ is always a terminal state since there is a single step to the POMDP.
    \item $R: S \times A \times S \rightarrow \left\{0, 1\right\}$ is the reward function defined as 
    \begin{equation}
        R(s, a, s') = \begin{cases}
            1 \text{ if } s' \text{ is classified as genuine}\\
            0 \text{ otherwise}
        \end{cases}
    \end{equation}
    \item $\Omega: S \rightarrow O$ is the observation function that extracts the observation from the state, i.e. extracts $x_k$ from $x$.
\end{itemize}

\subsection{RL agent}
We identify Proximal Policy Optimization~\cite[PPO]{SchulmanWDRK17-ppo} as a suitable single-agent Deep Reinforcement Learning algorithm for FRAUD-RLA due to its ability to handle continuous action spaces. Additionally, PPO is known to require little hyperparameter tuning compared to other Deep RL methods and has been proven to perform well in a wide variety of tasks \cite{schulman_proximal_2017, yu2022surprising}. 
The high-level working principle of FRAUD-RLA is shown in \autoref{alg:FRAUD-RLA}. 
At each round, FRAUD-RLA receives the fixed known features as input (line \ref{line:getfeatures}). It passes them to PPO, which generates a conditional multivariate Gaussian distribution over the controllable features space $R^C$ (line \ref{line:compute_meansCov}). FRAUD-RLA then samples from the distribution a set of controllable features $x_c$ (line \ref{line:bestControl}), passes them to the classifier, and receives the reward (line \ref{line:collect-rewardPPO}), measured as $1 - f(x_c, x_k, x_u)$, where  $x_u$ are the unknown features at this round. FRAUD-RLA then uses the reward and the fixed and controllable observations to train PPO (lines \ref{line:update1}-\ref{line:update2}). The mechanism repeats at each round.

Internally, PPO uses an actor-critic architecture~\cite{barto1983neuronlike}. The actor network (\autoref{tab:nn-actor}) takes as input the agent observation and outputs the action to take. The critic network (\autoref{tab:nn-critic}) takes the agent observations as input and outputs the observations' value, expressed as the sum of discounted rewards until the end of the episode. Since we are working in a single-step environment, we can express it as the reward associated with the observation.

The training process uses the reward as feedback to 
iteratively optimize the actor's policy using a clipped objective function to maximize the updates' stability.
It also optimizes the critic's value function using a mean squared error loss to minimize the difference between predicted and actual returns.
We show the structure of the actor and critic networks in \autoref{tab:nn-actor} and \autoref{tab:nn-critic}, respectively.
Since we want FRAUD-RLA to work in different settings and on different datasets without specific hyperparameter tuning, we opted for a straightforward architecture, with both networks presenting only two fully connected inner layers of $32$ nodes each. A normalization layer in the Critic Network was added to facilitate the algorithm convergence.

\begin{table}
    \centering
    \caption{Actor network architecture of PPO.}
    \label{tab:nn-actor}
    \begin{tabular}{|llr|}
         \hline
         \textbf{Layer type} & \textbf{Activation} & \textbf{Output size}  \\
         \hline
         Input & & $K$\\
         Linear & Tanh & $32$\\
         Linear & Tanh & $32$\\
         Linear &  & $C + C^2$\\
         \hline
    \end{tabular}
\end{table}

\begin{table}
    \centering
    \caption{Critic network architecture of PPO}
    \label{tab:nn-critic}
    \begin{tabular}{|llr|}
         \hline
         \textbf{Layer type} & \textbf{Activation} & \textbf{Output size}  \\
         \hline
         Input & & $K$\\
         LayerNorm & & $K$\\
         Linear & Tanh & $32$\\
         Linear & Tanh & $32$\\
         Linear &  & $1$\\
         \hline
    \end{tabular}
\end{table}


\subsubsection{Correlations in the action space}
Although the network architectures are similar to other works in the field of single-agent RL with continuous action spaces, we differ from most of them by learning the parameters (both the means and the covariance matrix) of a multivariate normal distribution for the action space. In contrast, most works in the domain~\cite{SchulmanWDRK17-ppo, duan_benchmarking_2016} only learn the means of a normal distribution and either use a hand-crafted constant or an annealed value for the variance.

Our choice to learn the covariance matrix is motivated by two reasons. First, we assume that the features of a transaction (i.e. the action) are correlated. For example, the fact that a terminal is located in a luxury store will influence other parameters such as the amount of the transaction and the type of credit card. Then, we assume that the attackers operate without prior knowledge of the data, forcing them to learn it throughout the training. The need to learn the covariance matrix is the reason why the Actor-Network has $C + C^2$ outputs (see \autoref{tab:nn-actor}) because $C$ values are used as the means and  $C^2$ values as the covariance matrix of the multivariate normal distribution from which actions are sampled.

\subsubsection{Partial observability}
Note that, unlike other works in the field of partially observable RL~\cite{hausknecht2015deep, sunehag2017value}, we do not use recurrent neural networks (RNN). The reason is that RNNs are typically used for an agent to remember what it has observed in the previous steps of an episode. Since we work in an environment where there is only one step in an episode, there is no need for recurrent networks.

\section{Experiments Design} \label{sec:Experiments Design}

In this section, we present an experimental analysis of our work. We first describe our experimental setup including our datasets, our baselines, and our methodology. 

\subsection{Experimental Setup}

\subsubsection{Datasets}
We evaluate our approach on three different datasets, each selected to represent different properties of fraud detection.
Because class balance does not directly affect FRAUD-RLA, all datasets have been balanced for these experiments. To guarantee Mimicry is not negatively affected by this change, we train it using only genuine transactions.
Our three datasets are as follows.

\begin{itemize}
    \item \textbf{Generator} Dataset generated employing the synthetic credit card fraud detection generator from~\cite{leborgne2022fraud}. The generator, used in previous works on fraud detection~\cite{lunghi2023adversary, paldino2024role}, is built on a set of customers interacting with terminals to create transactions, where frauds are inserted according to previously defined patterns. Contrary to other works, the resulting datasets maintain the features' semantics, allowing aggregated features to be composed.
    This allows us to divide features into purely transaction-based (like the amount), aggregated on the customer, and aggregated on the terminal. Depending on
    the threat model, the two types of aggregated features can be 
    controllable, known, and unknown features, respectively. 
    \item \textbf{Credit card fraud detection (Kaggle)} 
    Dataset~\cite{dal2015calibrating}
    obtain by applying a PCA transformation~\cite{MACKIEWICZ_1993} on 250K real transactions performed in September 2013 by European cardholders. It is one of the most widely used datasets in fraud detection. Each transaction has a label (genuine or fraudulent) and 30 features, including the amount. 
    Since the PCA transformation lost the semantics of the original columns, we randomly select features as known, controllable, and unknown. While not semantically meaningful, this allows us 
    to randomly assign a number of features as unknown and uncontrollable, measuring the impact of their number on attacks' performance.
    \item \textbf{SKLearn} A binary classification problem composed of clusters of normally distributed points generated using SKLearn~\cite{scikit-learn}. Although it, too, lacks the original feature semantics, it provides the possibility of selecting different class distributions and problem dimensionalities, which helps us generalize our findings.

\end{itemize}

Attackers may be unable to fine-tune their method before testing it against the classifier on a given dataset. For this reason, we will conduct all tests using the same PPO implementation and hyperparameters. While this is a pessimistic assumption for the attacker, it still provides a lower bound on the attack's effectiveness. Furthermore, this approach allows for a lower entrance barrier for deploying the attack, which, as discussed in Section \ref{sec:Threat}, makes the attack's potential consequences even more severe.

\subsubsection{Baseline: Mimicry} \label{sec:Mimicry}
As discussed in Section \ref{sec:Threat}, an attack that does not require access to a training set before it begins is generally easier to deploy. However, no attack reported in related literature can be employed against fraud detection engines under this condition.
Hence, to provide a fair baseline in our experiments, we assume that an attacker may access an unlabeled training set, which comprises an easier-to-meet condition than accessing the labeled one. Since data in fraud detection are typically highly skewed towards the genuine class~\cite{dal2015calibrating}, fraudsters may perform a Mimicry attack under this condition, therefore modeling genuine users' behavior.
Since FRAUD-RLA does not need or use such a training set, we effectively put FRAUD-RLA at a disadvantage as we compare it with attacks operating in a less challenging environment. Although this could potentially result into a pessimistic assessment of its relative effectiveness, still, it allows us to provide a reasonable baseline for FRAUD-RLA performance.
Should FRAUD-RLA remain competitive in such a setting, this would further showcase its effectiveness.


Mimicry has been employed to replicate the behavior of genuine users, including time-dependent features~\cite{carminati2018security}. However, in our case, we assume that attackers lack knowledge of time-dependent features. 
Therefore, we adopt a simplified version of Mimicry that approximates user behavior using straightforward statistical methods.
While more complex distributions may, in principle, be used, our approach has the significant advantage of requiring significantly less hyperparameter tuning, which aligns with the considered threat model. 

The resulting algorithm, illustrated in~\autoref{alg:Mimicry}, works as follows.
First, the attacker accesses a small training set and uses it to fit a statistical model (line \ref{line:fitMimicry}).
We test different data distributions: a uniform, various univariate normal distribution, a multivariate normal distribution, and a Gaussian mixture with $10$ mixture components. Concerning the training set size, we consider two cases: the realistic one, where the attacker can observe a small training set of $1000$ observations, and the pessimistic case, where we allow the attacker to observe the complete training set.
At each round, the attacker randomly samples the values of the controllable features from the trained distribution (line \ref{line:sampleGMimicry}) and sends them to the classifier. The resulting transaction combines the controllable features generated by Mimicry, as well as the fixed and the unknown features, and is evaluated by the classifier (line \ref{line:MimicryGetReward}).

\begin{algorithm}
\caption{Mimicry to maximize the sum of successful frauds (total reward) over training time}
\begin{algorithmic}[1]
\Require Time budget $t_{\max}$, Fraud Detection Engine $f$, Set of fraudulent transactions $X$, environment $M$, small training set $TR'$, distribution family $g$

\State $t \gets 0$
\State $total\_reward \gets 0$
\State Fit distribution $g \leftarrow g(TR')$ inside domain $ R^C$ \label{line:fitMimicry}

\While{$t < t_{\max}$}
    \State Receive $x_k, x_u$ from $M$ \label{line:getFeaturesMimicry}
    \State $x_c \leftarrow$ sample from $g$ \label{line:sampleGMimicry}
     \State $reward \gets  1 - f(x_c, x_k, x_u)$ \label{line:MimicryGetReward}
    \State $t \gets t + 1$
    \State $total\_reward \gets total\_reward + reward$
\EndWhile
\end{algorithmic}
\label{alg:Mimicry}
\end{algorithm}

\subsubsection{Methodology}

The first step of our evaluation is constructing our fraud detection engine. For this work, we imagine a simplified engine comprising two models: a machine-learning and a rule-based classifier. 
In line with previous works on credit card fraud detection~\cite{adewumi2017survey, dal2014learned}, we test two different machine learning classifiers: a Random Forest (RF) and a feed-forward neural network (NN). Both algorithms are trained with a standard random grid cross-validation strategy for hyperparameters tuning~\cite{scikit-learn}. 
To penalize strategies based on extreme values, we pair the machine learning model with a rule-based classifier rejecting transactions where any feature would fall into the $10\%$ most extreme values in the training set.
Finally, we run the attacks and measure their improvement over time. We use the average success rate to evaluate the attacks' success, calculated as the percentage of successful frauds. In particular, we measure it over the first $300$, $1000$, and $4000$ frauds.

\subsection{Experimental Findings} \label{sec:results}

Next, we present our experimental analysis first using the synthetic data generator, and then, using real data and the SKLearn generator.

\begin{table}
\centering
\label{tab:classification}
\begin{tabular}{|l|l|l|l|l|}
\hline
 & Accuracy & Precision & Recall & F1 \\
 \hline
 Generator: RF & 0.89 & 0.95 & 0.82 & 0.88 \\ 
\hline
Generator: NN & 0.89 & 0.95 & 0.81 & 0.88 \\ 
\hline
Kaggle: RF & 0.95 & 1.00 & 0.91 & 0.95 \\ 
\hline
Kaggle: NN & 0.93 & 0.97 & 0.90 & 0.93 \\ 
\hline

SKLearn 0: RF & 1.00 & 1.00 & 1.00 & 1.00 \\ 
\hline
SKLearn 0: NN & 1.00 & 1.00 & 1.00 & 1.00 \\ 
\hline

SKLearn 1: RF & 0.99 & 0.99 & 0.99 & 0.99 \\ 
\hline
SKLearn 1: NN & 0.99 & 0.99 & 0.99 &0.99 \\ 
\hline

SKLearn 2: RF & 0.74 & 0.75 & 0.74 & 0.75 \\ 
\hline
SKLearn 2: NN & 0.82 & 0.82 & 0.83 & 0.82 \\ 
\hline
\end{tabular}
\caption{Classifiers Performance over the tested datasets.}
\label{tab:Classifiers}
\end{table}

\paragraph{Synthetic data generator.}

\begin{table*}[ht]
\centering
\resizebox{0.7\textwidth}{!}{%
\begin{tabular}{|c c||r|r|r|r|r|r|r|r|}
\hline 
\multicolumn{2}{|c||}{\textbf{Features} } & \multicolumn{8}{c|}{\textbf{Baseline Algorithms}}     \\

\multicolumn{2}{|c||}{} & 
\multicolumn{2}{c|}{\textbf{Multivar}} &  
\multicolumn{2}{c|}{\textbf{Univar}} &
\multicolumn{2}{c|}{\textbf{Uniform}} &
\multicolumn{2}{c|}{\textbf{Mixture}}\\

\textbf{Fixed} & \textbf{Unknown} &
 \textbf{1k} & \textbf{100\%} & \textbf{1k} & \textbf{ 100\%} & \textbf{1k} & \textbf{100\%} & \textbf{1k} & \textbf{100\%} \\
 
\hline \hline
 / & / & 0.84 & 0.83 & 0.67 & 0.66 & 0.68 & 0.69 & 0.85 & 0.86 \\
\hline
 / & T & 0.52 & 0.52 & 0.40 & 0.39 & 0.39 & 0.39 & 0.57 & 0.57 \\
\hline
 / & C & 0.60 & 0.60 & 0.60 & 0.60 & 0.66 & 0.65 & 0.56 & 0.56 \\
\hline
 / & C, T & 0.23 & 0.23 & 0.23 & 0.23 & 0.26 & 0.26 & 0.23 & 0.23 \\
\hline
 T & / & 0.52 & 0.52 & 0.40 & 0.39 & 0.39 & 0.39 & 0.57 & 0.58 \\
\hline
T & C & 0.23 & 0.23 & 0.23 & 0.23 & 0.26 & 0.26 & 0.23 & 0.23 \\
\hline
C & / & 0.60 & 0.60 & 0.60 & 0.60 & 0.66 & 0.65 & 0.56 & 0.56 \\
\hline
C & T & 0.23 & 0.23 & 0.23 & 0.23 & 0.26 & 0.26 & 0.23 & 0.23 \\
\hline
C, T & / & 0.23 & 0.23 & 0.23 & 0.23 & 0.26 & 0.26 & 0.23 & 0.23 \\
\hline
\end{tabular}%
}
\caption{Generator, Random Forest. Baselines Success Rate comparison under various scenarios. Each baseline is trained in two ways: on $1000$ observations and over the whole training set.}
\label{tab:baselines_RF_Gen}
\end{table*}

\begin{table*}[ht]
\centering
\resizebox{0.7\textwidth}{!}{%
\begin{tabular}{|l l||r|r|r|r|r|r|r|r|}
\hline
\multicolumn{2}{|c||}{\textbf{Features} } & \multicolumn{8}{c|}{\textbf{Baseline Algorithms}}    
\\
\multicolumn{2}{|c||}{} & 
\multicolumn{2}{c|}{\textbf{Multivar}} &  
\multicolumn{2}{c|}{\textbf{Univar}} &
\multicolumn{2}{c|}{\textbf{Uniform}} &
\multicolumn{2}{c|}{\textbf{Mixture}}\\

\textbf{Fixed} & \textbf{Unknown} &
  \textbf{1k} & \textbf{100\%} & \textbf{1k} & \textbf{ 100\%} & \textbf{1k} & \textbf{100\%} & \textbf{1k} & \textbf{100\%} \\
\hline \hline

 / & / & 0.90 & 0.89 & 0.72 & 0.71 & 0.67 & 0.66 & 0.89 & 0.89 \\
\hline
 / & T & 0.58 & 0.58 & 0.45 & 0.44 & 0.43 & 0.43 & 0.60 & 0.60 \\
\hline
/ & C & 0.69 & 0.69 & 0.70 & 0.69 & 0.75 & 0.72 & 0.66 & 0.66 \\
\hline
 / & C, T  & 0.32 & 0.32 & 0.32 & 0.31 & 0.33 & 0.33 & 0.32 & 0.32 \\
\hline
T & / & 0.58 & 0.57 & 0.45 & 0.44 & 0.43 & 0.43 & 0.60 & 0.60 \\
\hline
T & C & 0.32 & 0.32 & 0.32 & 0.32 & 0.33 & 0.33 & 0.31 & 0.32 \\
\hline
C & /  & 0.70 & 0.69 & 0.70 & 0.69 & 0.75 & 0.72 & 0.66 & 0.66 \\
\hline
C & T & 0.32 & 0.32 & 0.32 & 0.32 & 0.33 & 0.33 & 0.31 & 0.32 \\
\hline
C, T & / & 0.32 & 0.32 & 0.32 & 0.32 & 0.33 & 0.33 & 0.31 & 0.32 \\
\hline
\end{tabular}%
}
\caption{Generator, Neural Network. Baselines Success Rate comparison under various scenarios. Each baseline is trained in two ways: on $1000$ observations and over the whole training set. }
\label{tab:baselines_NN_Gen}
\end{table*}

First, we analyze the Mimicry results, which provide a baseline for attackers' performance. 

To do so, we first compare the performance of the different Mimicry techniques we used. 
We show the analysis results in
\autoref{tab:baselines_RF_Gen} and \autoref{tab:baselines_NN_Gen}, where we measure the attacks' success rate over different settings, where each setting is defined by the types of fixed, known features (\textit{FIXED} in the tables) and those unknown to the attacker (\textit{UNKNOWN}).
We group here features as terminal-based (\textit{T}) and customer-based (\textit{C}) aggregations.
For datasets without feature semantics, such as Kaggle and SKLearn, we will instead use the percentage of features belonging to each group.
This table, comparing the performance of multiple baselines trained over a dataset of $1000$ observations (\textit{1K}) or over the full Training Set (\textit{100\%}),
allows us to make some considerations.

First, increasing the size of the training set over $1000$ samples does not significantly improve the performance of the algorithms. Second, while the recall of Random Forest and Neural Network on the test is set is practically the same (as shown in \autoref{tab:Classifiers}), all Mimicry attacks achieve a significantly higher success rate against Neural Networks, showing the superior robustness of Random Forests. This aligns with previous findings in the literature, where Random Forests proved to be more robust than deep learning algorithms against traditional adversarial attacks~\cite{ding2019defending}. Finally, when fraudsters control all the features, Mimicry attacks achieve a very high success rate against classifiers having recalls of over $0.9$ on non-adversarial data. However, attacks' effectiveness dramatically decreases when we reduce the number of controllable features, as shown by the decreasing success rate.  

Next, we 
compare FRAUD-RLA with the best-performing baseline under all settings.
We show the results in  \autoref{tab:Generator-SuccessRate-RF} and \autoref{tab:Generator-SuccessRate-NN}. Here, for each setting, we compare the best baseline (\textit{Best Baseline} in the table) with the average success of FRAUD-RLA after $300$, $1000$ and $4000$ attacks, respectively.
Both tables show an increase in the success rate over time as FRAUD-RLA learns the best attack policy. In the case of random forests, FRAUD-RLA starts with an average cumulative reward 
that is significantly lower than the baseline but increases until it reaches or surpasses that value. Neural Networks instead are breached from the first rounds by FRUAD-RLA, confirming their inferior robustness.
Above all, FRAUD-RLA in most settings surpasses the baselines in Average Cumulative Reward from early rounds, optimizing the exploration-exploitation tradeoff. 
It should be noted that, in the case of this dataset, knowing fixed features without being able to control them does not help FRAUD-RLA improve its performance, as customers' and terminals' features are created by independent processes in the generator~\cite{leborgne2022fraud}.

\begin{table}
\resizebox{\columnwidth}{!}{
\begin{tabular}{|l l||c c c||c|}
\hline
\multicolumn{2}{|c||}{\textbf{Features} } & \multicolumn{3}{c||}{\textbf{FRAUD-RLA Time}}     & \multicolumn{1}{c|}{\textbf{Best  } } \\
\textbf{Fixed} & \textbf{Unknown} & \textbf{300} & \textbf{1000} & \textbf{4000} &{\textbf{Baseline}}  \\
\midrule
/ & / & 0.15 & 0.37 & 0.75 & 0.86 \\
/ & T & 0.12 & 0.28 & 0.53 & 0.57 \\
/ & C & 0.32 & 0.57 & 0.74 & 0.66 \\
/ & C, T& 0.24 & 0.25 & 0.28 & 0.26 \\
T & / & 0.13 & 0.30 & 0.53 & 0.58 \\
T & C & 0.24 & 0.25 & 0.28 & 0.26 \\
C & / & 0.33 & 0.57 & 0.73 & 0.66 \\
C & T & 0.24 & 0.25 & 0.28 & 0.26 \\
C, T& / & 0.24 & 0.25 & 0.28 & 0.26 \\
\bottomrule
\end{tabular}
}
\caption{Generator, Random Forest. Comparison between the average Cumulative Reward of FRAUD-RLA over $300$, $1000$ and $4000$ frauds with the best baseline in each setting.}
\label{tab:Generator-SuccessRate-RF}
\end{table}

\begin{table}
\resizebox{\columnwidth}{!}{%
\begin{tabular}{|l l||c c c||c|}
\hline
\multicolumn{2}{|c||}{\textbf{Features} } & \multicolumn{3}{c||}{\textbf{FRAUD-RLA Time}}     & \multicolumn{1}{c|}{\textbf{ Best} } \\
\textbf{Fixed} & \textbf{Unknown} & \textbf{300} & \textbf{1000} & \textbf{4000} & {\textbf{Baseline}}  \\
\midrule
/ & / & 0.72 & 0.86 & 0.89 & 0.90 \\
/ & T & 0.64 & 0.85 & 0.91 & 0.60 \\
/ & C & 0.79 & 0.90 & 0.92 & 0.75 \\
/ & C T & 0.80 & 0.93 & 0.97 & 0.33 \\
T & / & 0.62 & 0.85 & 0.91 & 0.60 \\
T & C & 0.77 & 0.92 & 0.97 & 0.33 \\
C & / & 0.79 & 0.91 & 0.93 & 0.75 \\
C & T & 0.78 & 0.92 & 0.97 & 0.33 \\
C, T & / & 0.79 & 0.93 & 0.97 & 0.33 \\
\bottomrule
\end{tabular}%
}
\caption{Generator, Neural Network. Comparison between the average Cumulative Reward of FRAUD-RLA over $300$, $1000$ and $4000$ frauds with the best baseline in each setting.}
\label{tab:Generator-SuccessRate-NN}
\end{table}

\begin{table}[ht]
\resizebox{\columnwidth}{!}{
\begin{tabular}{|ll ||ccc||c|}
\hline
\multicolumn{2}{|c||}{\textbf{Features} } & \multicolumn{3}{c||}{\textbf{FRAUD-RLA Time}}     & \multicolumn{1}{c|}{\textbf{Best} } \\
\textbf{Fixed} & \textbf{Unknown} & \textbf{300} & \textbf{1000} & \textbf{4000} & {\textbf{Baseline}}  \\
\midrule
0\%  & 0\%  & 0.87 & 0.90 & 0.91 & 0.99 \\
0\%  & 25\% & 0.83 & 0.89 & 0.90 & 0.91 \\
25\% & 0\%  & 0.86 & 0.91 & 0.91 & 0.92 \\
0\%  & 50\% & 0.76 & 0.86 & 0.89 & 0.71 \\
25\% & 25\% & 0.77 & 0.88 & 0.90 & 0.71 \\
50\% & 0\%  & 0.76 & 0.87 & 0.90 & 0.65 \\
25\% & 50\% & 0.46 & 0.57 & 0.64 & 0.34 \\
50\% & 25\% & 0.48 & 0.62 & 0.69 & 0.35 \\
\bottomrule
\end{tabular}
}
\caption{Kaggle, Random Forest. Comparison between the average Cumulative Reward of FRAUD-RLA over $300$, $1000$ and $4000$ frauds with the best baseline in each setting.}
\label{tab:Kaggle-SuccessRate-RF}
\end{table}

\paragraph{Real data and SKLearn generator.}

We continue our 
analysis over one real and various synthetic datasets, starting with the real dataset, named in the experiments the Kaggle Dataset. 

\autoref{tab:Classifiers} shows that the classifiers perform better on this dataset than on the generator data, with recall values from $0.9$ for both RF and NN.
Looking at the cumulative success rate, we see in \autoref{tab:Kaggle-SuccessRate-RF} and \autoref{tab:Kaggle-SuccessRate-NN} how the baseline is still highly effective when having full control over the features. However, this quickly decreases with the number of fixed and unknown features. Instead, FRAUD-RLA average success rate decreases very slowly with the number of fixed or unknown features, with only some configuration of Random Forest leading to a success rate below $0.9$. Interestingly, in this dataset, known fixed features seem to improve the performance of FRAUD-RLA compared to unknown features. More tests should be performed to study the significance of this effect.

Finally, we use SKLearn to study the impact of dimensionality and task complexity on the attack's effectiveness. 
To do so, we focus on three specific settings: 
\begin{itemize}
    \item Scenario 1 (easy): A problem with $16$ features, $1$ cluster per class, and high separation between the classes.
    
    \item Scenario 2 (intermediate): A problem with $64$ features, $8$ cluster per, class and high separation between the classes.
    
    \item Scenario 3 (hard): A problem with $64$ features, $16$ cluster per class, and small separation between the classes.

\end{itemize}
\begin{table}
\resizebox{\columnwidth}{!}{%
\begin{tabular}{|ll ||ccc||c|}
\hline
\multicolumn{2}{|c||}{\textbf{Features} } & \multicolumn{3}{c||}{\textbf{FRAUD-RLA Time}}     & \multicolumn{1}{c|}{\textbf{ Best} } \\
\textbf{Fixed} & \textbf{Unknown} & \textbf{300} & \textbf{1000} & \textbf{4000} & {\textbf{Baseline}}  \\
\midrule
0\%  & 0\%  & 0.87 & 0.90 & 0.93 & 0.94 \\
0\%  & 25\% & 0.87 & 0.91 & 0.94 & 0.58 \\
25\% & 0\%  & 0.86 & 0.92 & 0.94 & 0.60 \\
0\%  & 50\% & 0.88 & 0.92 & 0.95 & 0.41 \\
25\%& 25\% & 0.88 & 0.93 & 0.95 & 0.39 \\
50\% & 0\%  & 0.87 & 0.92 & 0.95 & 0.41 \\
25\% & 50\% & 0.88 & 0.94 & 0.96 & 0.26 \\
50\% & 25\% & 0.88 & 0.94 & 0.96 & 0.28 \\
\bottomrule
\end{tabular}%
}
\caption{Kaggle, Neural Network. Comparison between the average Cumulative Reward of FRAUD-RLA over $300$, $1000$ and $4000$ frauds with the best baseline in each setting.}
\label{tab:Kaggle-SuccessRate-NN}
\end{table}

\begin{table*}[]
\centering
\resizebox{0.8\textwidth}{!}{%
\begin{tabular}{|ll||lll|lll|}
\hline
\multicolumn{2}{|c||}{\textbf{Features} } & \multicolumn{3}{c|}{\textbf{Best Baseline}}     & \multicolumn{3}{c|}{\textbf{FRAUD-RLA Variation from baseline} } \\
\textbf{Fixed}        & \textbf{Unknown}       & \textbf{Dataset 1} &\textbf{Dataset 2} & \textbf{Dataset 3} &\textbf{Dataset 1}  & \textbf{Dataset 2} & \textbf{Dataset 3} \\ \hline
0 \%           & 0 \%             & 1         & 1         & 0.72      & -0.12      & -0.47     & 0.02      \\
0 \%           & 25\%             & 0.97      & 0.93      & 0.62      & -0.12      & -0.41     & 0.17      \\
25\%           & 0\%              & 0.99      & 0.94      & 0.63      & -0.11      & -0.43     & 0.09      \\
0 \%           & 50\%             & 0.53      & 0.45      & 0.5       & 0.08       & -0.05     & 0.13      \\
25\%           & 25\%             & 0.43      & 0.46      & 0.5       & 0.17       & -0.04     & 0.12      \\
50\%           & 0\%              & 0.58      & 0.46      & 0.5       & 0.05       & 0.12      & 0.10      \\
25\%           & 50\%             & 0.03      & 0.04      & 0.37      & 0.00       & 0.03      & 0.24      \\
50\%           & 25\%             & 0.01      & 0.04      & 0.38      & 0.00       & 0.02      & 0.23      \\ \hline
\end{tabular}%
}
\caption{SKLearn, Random Forest. Best baseline performance over different datasets and FRAUD-RLA improvement over it measured after $4000$ frauds.}
\label{tab:SKLearn-SuccessRate-RF}
\end{table*}

\begin{table*}[]
\centering
\resizebox{0.8\textwidth}{!}{%
\begin{tabular}{|ll||lll|lll|}
\hline
\multicolumn{2}{|c||}{\textbf{Features}} & \multicolumn{3}{c|}{\textbf{Baseline}}     & \multicolumn{3}{c|}{\textbf{FRAUD-RLA Variation from baseline}} \\
\textbf{Fixed}       & \textbf{Unknown}      & \textbf{Dataset 1} & \textbf{Dataset 2} & \textbf{Dataset 3} & \textbf{Dataset 1}      & \textbf{Dataset 2}      & \textbf{Dataset 3}      \\ \hline
0\%   & 0\%   & 1.00 & 1.00 & 0.80 & -0.08 & -0.22 & -0.07 \\
0\%   & 25\%  & 0.91 & 0.87 & 0.65 & 0.02  & -0.06 & 0.10  \\
25\%  & 0\%   & 0.97 & 0.87 & 0.66 & 0.04  & -0.06 & 0.10  \\
0 \%  & 50\%  & 0.39 & 0.44 & 0.49 & 0.54  & 0.39  & 0.30  \\
25\%  & 25\%  & 0.42 & 0.45 & 0.50 & 0.51  & 0.39  & 0.31  \\
50\%  & 0\%   & 0.45 & 0.41 & 0.49 & 0.49  & 0.43  & 0.31  \\
25\%  & 50\%  & 0.02 & 0.08 & 0.33 & 0.94  & 0.80  & 0.54  \\
50\%  & 25\%  & 0.04 & 0.09 & 0.33 & 0.92  & 0.75  & 0.55  \\ \hline
\end{tabular}%
}
\caption{SKLearn, Neural Network. Best baseline performance over different datasets and FRAUD-RLA improvement over it measured after $4000$ frauds.}
\label{tab:SKLearn-SuccessRate-NN}
\end{table*}


\autoref{tab:Classifiers} illustrates that both classifiers perform nicely when applied in the first two scenarios while struggling in the third one. 
We show the best baseline performance and how FRAUD-RLA compares with it in  \autoref{tab:SKLearn-SuccessRate-RF} and \autoref{tab:SKLearn-SuccessRate-NN}. In both tables, we only report the variation of FRAUD-RLA from the baseline after $4000$ attacks. 

Similarly to the other datasets, Neural Networks, despite being more powerful than random forests on the original datasets, are more frequently breached by FRAUD-RLA, confirming their inferior robustness.
Concerning the attacks, FRAUD-RLA is the best attack against Neural Networks, where it achieves an excellent success rate in all scenarios, particularly compared to Mimicry.
Experiments conducted on Random Forests give a more complex picture. FRAUD-RLA is still the best method in Scenario $3$, with attacks breaking the classifier even when most features are uncontrollable or unknown. Scenarios $1$ and $2$ are difficult for all attacks, but in Scenario $2$, the baseline outperforms FRAUD-RLA. 
Investigating this in greater detail, it turns out that the current implementation of FRAUD-RLA performs better in exploration scenarios, and hence, it outperforms Mimicry when more features are hidden. Instead, Mimicry thrives in a setting where classes are clearly divided and all features are controllable. This is more evident in the first three lines where Mimicry performs better than FRAUD-RLA.
It is worth noting that such cases, although useful in a thorough analysis, are very unlikely in real-world settings where most features are routinely unknown or fixed. Hence, we expect FRAUD-RLA to be an excellent tool for most practical scenarios.

\section{Conclusions and future work} \label{sec:Conclusions}

Adversarial attacks are a growing threat, and credit card fraud detection systems are sensitive targets.
The lack of research at the intersection of the two domains is a potential vulnerability for existing fraud detection engines. It increases the likelihood of catastrophic consequences should fraudsters identify and implement an effective attack strategy. 
This work aimed to mitigate this problem by analyzing the domain's main features, focusing on differences with more traditional domains, such as image recognition and malware detection.

Understanding the challenges attackers face is crucial when designing and modeling the threat of advanced machine learning for fraud detection. Therefore, we expanded existing threat models to adapt them to credit card fraud detection. Specifically, we updated existing approaches to consider the lack of human supervision, the limited data knowledge and control, and the exploration-exploitation tradeoff.
Using the defined threat model as a blueprint, we developed FRAUD-RLA, a novel attack designed explicitly to tackle the aforementioned issues. To do so, we modeled the problem of finding a successful fraudulent pattern in the shortest possible time as a Partially Observable Markov Decision Process, where the known fixed features represent the visible state and the action corresponds to selecting the optimal values for the controllable features. To optimize this problem, we employed Proximal Policy Optimization (PPO), a robust algorithm able to optimize continuous policies with little hyperparameter optimization.
Our experiments show that FRAUD-RLA quickly achieves a high average success rate under most settings, consistently beating the baselines without optimizing the hyperparameters against the different datasets and target classifiers.

It is worth noting that we do not aiming at developing an "off-the-shelf" attack that could be directly applied in a real-world scenario.
The presence of categorical variables, eventual limitations to the frequency at which frauds can be performed without raising any alarm, and, in general, the need to adapt the attack to various challenges that may arise in the real case, all serve to illustrate why FRAUD-RLA does not constitute an immediate threat to any real-world system, and is not, as such, a valuable tool for malignous agents.
On the contrary, improving our understanding of adversarial security of fraud detection systems will be crucial in developing effective defenses. As security is frequently evaluated through the lens of red teaming~\cite{avin2021filling}, it is imperative to use the appropriate tools for comprehensive assessments and improvements, and we designed FRAUD-RLA to fill that role eventually.


Future work may span several challenging directions. 
First, different reinforcement learning algorithms could be tried, such as contextual bandit algorithms~\cite{pmlr-v151-kassraie22a, zhou_neural_2020}, that are dedicated to stateful single-step problems. Moreover, FRAUD-RLA could be extended to tackle other fraud detection aspects that we did not consider in this work, such as the presence of categorical variables and the possibility of delayed feedback caused by the presence of humans in the loop.
More generally, assessing FRAUD-RLA in even more realistic environments is a necessary step to move over the lab-only evaluation phase~\cite{arp2022and} and transform it into a proper system to evaluate the robustness of the existing engines in production.
Another promising direction would be to use FRAUD-RLA to develop solid defenses that can limit the effectiveness of attacks based on reinforcement learning against credit card fraud detection.
Following this direction, we are currently researching the 
effectiveness of training a classifier to prioritize learning from features that are uncontrollable or unknown to the attacker to achieve \textit{robust by design} credit card fraud detection.

\section{Ethics Considerations}
Machine learning systems are employed in various applications, and research should always make them more secure. Unresponsable disclosure of vulnerabilities can give attackers a head start and severely limit the systems' security. Researching vulnerabilities, however, remains crucial to mitigate the risk posed by zero-day threats. 
Adversarial machine learning literature, for instance, is built on the assumption that published attacks improve our understanding of machine learning models' vulnerabilities.

In this work, we did not aim to exploit vulnerabilities of any specific system. Instead, we used open research as a baseline to define how fraud detection engines work. Our threat model and the attack we designed are not intended to directly apply to any system in production. As stated in Section \ref{sec:Conclusions}, this would require updating the attack to tackle issues like categorical variables, transaction frequency caps, and, in general, multiple challenges real-world systems present that are not described in the literature.
However, reinforcement learning does threaten fraud detection, as it has been shown to do in malware detection~\cite{anderson2018learning} and as we showed in this work. In reality, the lack of research on the topic is arguably one of the main vulnerabilities, as it increases the chance attackers may find successful strategies that employ reinforcement learning before the community has a solid understanding of the threat.

Therefore, this work's primary goal is to provide a framework to study these attacks and how they can challenge systems' robustness. First, this could be used by fraud detection practitioners to evaluate their systems, possibly extending and modifying FRAUD-RLA to use it for their own tests. Ultimately, however, our work is intended to help us understand how to defend against RL-based attacks. Hopefully, FRAUD-RLA will help us and other researchers towards achieving this goal in the shortest possible time.
Finally, all experiments were conducted on openly available data without targeting any in-production system. Therefore, no participant was harmed in the writing of this paper nor in the design and test of the attack.

\section{Open Science}
The threat model and the theoretical analysis were based on published works, and this paper is designed to be fully open and reproducible.
Specifically, the Generator code is available at \cite{leborgne2022fraud}, the Kaggle Dataset at  \href{https://www.kaggle.com/datasets/mlg-ulb/creditcardfraud}{Kaggle Dataset}, and SKLearn is an open library. The code of the experiments will be released and openly available on Github, toghether with the instructions to set up the datasets and the experiments.
Finally, the experiments were designed to be open and fair. Metrics, hyperparameters, and datasets were chosen for their applicability to credit card fraud detection and are coherent with previous works and the problem.


\bibliographystyle{plain}
\bibliography{biblio}

\newpage

\end{document}